\relax
\documentclass[twocolumn ,11pt]{article} 
\usepackage{times}  
\usepackage{helvet} 
\usepackage{courier}  
\usepackage[hyphens]{url}  
\usepackage{graphicx}  
\usepackage{appendix}

\usepackage{algorithm}
\usepackage{algorithmic}
\usepackage{amsmath}
\usepackage{mathtools}
\usepackage{amssymb}
\usepackage{latexsym}
\usepackage{xr}
\usepackage{subcaption}
\usepackage{csvsimple}

\usepackage{times,fullpage}

\newcommand{\fl}{\textit{FL}}
\newcommand{\flinst}{\textit{fl}}

\newcommand{\decoder}{\textit{d}}
\newcommand{\encoder}{\textit{e}}
\DeclareMathOperator*{\argmax}{arg\,max}
\newcommand{\dom}{\textit{dom}} 

\newtheorem{THEOREM}{Theoremmy}

\newtheorem{COROLLARY}[THEOREM]{Corollary}

\newtheorem{PROPOSITION}[THEOREM]{Proposition}

\newtheorem{DEFINITION}[THEOREM]{Definition}
\newenvironment{definition}{\begin{DEFINITION} \hspace{-.85em} {\bf :} \rm}
                            {\end{DEFINITION}}
\newtheorem{CLAIM}[THEOREM]{Claim}

\newtheorem{EXAMPLE}[THEOREM]{Example}

\newtheorem{REMARK}[THEOREM]{Remark}

							\newtheorem{NOTATION}[THEOREM]{Notation}

\input epsf

\begin{document}

\title{Logical Interpretations of Autoencoders}

\author{Anton Fuxjaeger\\
University of Edinburgh\\ \texttt{anton.fuxjaeger@ed.ac.uk}
\and Vaishak Belle\\
University of Edinburgh \& Alan Turing Institute\\ \texttt{vaishak@ed.ac.uk} }

\date{}

\maketitle

\begin{abstract}
The unification of low-level perception and high-level reasoning is a long-standing problem in artificial intelligence, which has the potential to not only bring the areas of logic and learning closer together but also demonstrate how abstract concepts might emerge from sensory data.  Precisely because deep learning methods dominate perception-based learning, including vision, speech, and linguistic grammar, there is fast-growing literature on how to integrate symbolic reasoning and deep learning. Broadly, efforts seem to fall into three camps: those focused on defining a logic whose formulas capture deep learning, ones that integrate symbolic constraints in deep learning, and others that allow neural computations and symbolic reasoning to co-exist separately, to enjoy the strengths of both worlds. In this paper, we identify another dimension to this inquiry: what do the hidden layers really capture, and how can we reason about that logically? In particular, we consider autoencoders that are widely used for dimensionality reduction and inject a symbolic generative framework onto the feature layer. This allows us, among other things, to generate example images for a class to get a sense of what was learned. Moreover, the modular structure of the proposed model makes it possible to learn relations over multiple images at a time, as well as handle noisy labels. Our empirical evaluations show the promise of this inquiry.
\end{abstract}

\section{Introduction}
\noindent The unification of low-level perception and high-level reasoning is a long-standing problem in artificial intelligence, which has the potential to not only bring the areas of logic and learning closer together but also demonstrate how abstract concepts might emerge from sensory data. Precisely because deep learning methods dominate perception-based learning, including vision, speech, and linguistic grammar, there is fast-growing literature on how to integrate symbolic reasoning and deep learning. Efforts have ranged from providing a truth-theory to deep learning ~\cite{serafini2016logic,garcez2012neural}, neural architectures that enable differential computation for symbolic constraints~\cite{semanticLossXu2018,bovsnjak2017programming,rocktaschel2017end,santoro2017simple},
and embeddings for graph and relational data~\cite{yang2014joint,lin2015learning,niepert2016learning,dumancic2018auto}. 
Approaches such as DeepProbLog~\cite{manhaeve2018deepproblog}, on the other hand, treat deep learning as an external computation and integrate its predictions as an external predicate in a probabilistic logic programming framework. Broadly, efforts seem to fall into three camps: those focused on \textit{semantic characterizations} (i.e., define a logic whose formulas capture deep learning), \textit{constrained learning} (i.e., integrate symbolic constraints in deep learning), and \textit{hybrid methods} (allow neural computations and symbolic reasoning to co-exist separately, to enjoy the strengths of both worlds). 

In this paper, we identify another dimension to this inquiry: \textit{what do the hidden layers really capture, and how can we reason about that logically?} In particular, we consider autoencoders (AEs)~\cite{goodfellow2016deep,kingma2013auto,rezende2014stochastic}. As a variant of neural networks, AE frameworks are perhaps the most popular for dimensionality reduction, but its inner workings are entirely opaque and mysterious. Basically, given an encoder \encoder, one first applies it to input data $x$ to obtain a feature layer (\fl) and then attempts to recover $x$ from \fl~using a decoder \decoder. Constraints on \fl~can lead to massive reductions on the dimensionality and identify salient features for applications such as anomaly detection. (See, for example,~\cite{dumancic2018auto} that is a purely logical approach inspired by autoencoding principles.) Thus, we ask the question: \textit{can we inject a logical language onto the \fl~to perform Boolean reasoning over the \fl's variables?}

The exact choice of the language would depend on what we intend to do with the logic. A purely discrete representation such as propositional logic may not be very interesting or insightful about what the \fl~really captures, especially in cases where there may be probabilities assigned to image labels. In that regard, there has been an interesting development in knowledge representation over the last few years. As a special case of probabilistic logical models~\cite{de2015probabilistic,getoor2007statistical} tractable probabilistic models have emerged as an extension to data structures such as binary decision diagrams (BDDs). In particular probabilistic sentential decision diagram (PSDDs)~\cite{kisa2014probabilistic}, for example, are a complete and canonical representation of a probabilistic distribution defined over the models of a propositional theory. By imposing certain properties on the propositional representation, such as decomposability and determinism, probabilistic queries can be answered in polynomial time in the size of the data structure by way of model counting. Its parameters can be learned efficiently from data, which allows us to view the representation through a generative lens over a logical base. We discuss below how these features are put to use, but more generally, we see the work as a step in re-purposing deep learning in logical space to contributing to the emergence of high-level reasoning from a low-level system. Our contributions are orthogonal in many regards to the existing literature on neuro-symbolic systems and thus we imagine there would be space for looking at other kinds of integration with the existing literature.

Interestingly, we take note of multiple approaches for visually inspecting and interpreting NNs in the literature~\cite{szegedy2013intriguing,yosinski2015understanding}, with a special focus on understanding convolutions of deep networks after training~\cite{simonyan2013deep,zeiler2014visualizing}. While many of these methods yield various analysis of what happens in a given NN, including saliency maps~\cite{simonyan2013deep}, they differ in thrust significantly from our contributions. For example, optimization methods are usually used to infer and decode regions of interest in a specific layer of a pre-trained network. In contrast to this, our logical approach uses a symbolic framework to make sense of the NN over a generative model. As such we do not infer the meaning of individual variables, although it is possible to visualize them, but rather compute conditional probabilities over those variables.

It should also be noted that recent circuit models attempt to  tackle vision problems too (e.g., ~\cite{poon2011sum,gens2012discriminative,liang2019learning}), and so it is possible to realize the entire image classification pipeline using these tractable probabilistic models (however usually as a discriminative model). We think this is a very exciting development. What our work attempts to do, however, is to inspect state-of-the-art deep learning architectures (especially models like AEs that are very powerful for dimensionality reduction) via such symbolic generative models, in case such architectures are already in place, or are tackling problems still to be addressed using a pure circuit scheme. In that regard, as mentioned, our work is to be seen as attempting to re-purpose the latent space in a logical manner.

Our approach offers the following capabilities. We learn a PSDD over a discretized \fl, which yields a joint distribution over the individual variables, including image labels, of the \fl. This allows us, among other things, to visualize these individual variables by conditional sampling. In particular, this enables us to generate example images for a class to get a sense of what was learned. Moreover, the modular structure of the proposed model makes it possible to learn relations over multiple images at a time. Finally, because of the logical structure that we impose, noisy labels can also be handled.
We also discuss how we can evaluate the learned representation over well-known datasets, and also discuss both reconstructability (i.e., generative capabilities) and classification accuracy. 

At the outset, from an engineering (as opposed to mathematical) viewpoint, 
it should be noted that, at this point, since circuit software packages have not enjoyed the same amount of maturity as deep learning packages,  the reported accuracy is not as competitive as state-of-the-art systems. Nonetheless, although 
this reported accuracy is lower, the model has considerably more functionality, including the ability to sample prototypical images for each of the learned classes, ultimately aiding us in understanding what has been learned by the model (i.e., visually showing us what the model thinks a given class represents).

\section{Preliminaries}

\subsection{Probabilistic Sentential Decision Diagrams}
\label{sec:psdd}

Sentential decision diagrams (SDDs) were first introduced in ~\cite{darwiche2011sdd} and are tractable representations of propositional knowledge bases. SDDs are shown to be a strict subset of deterministic decomposable negation normal form (d-DNNF), a popular representation for probabilistic reasoning applications ~\cite{chavira2008probabilistic} due to their desirable properties. Decomposability and determinism especially ensure tractable probabilistic inference. PSDDs extend SDDs with probabilities, and are a complete and canonical representation of joint probability distributions~\cite{kisa2014probabilistic}.

\begin{figure}
\centering
\centerline{\includegraphics[width=.35\textwidth]{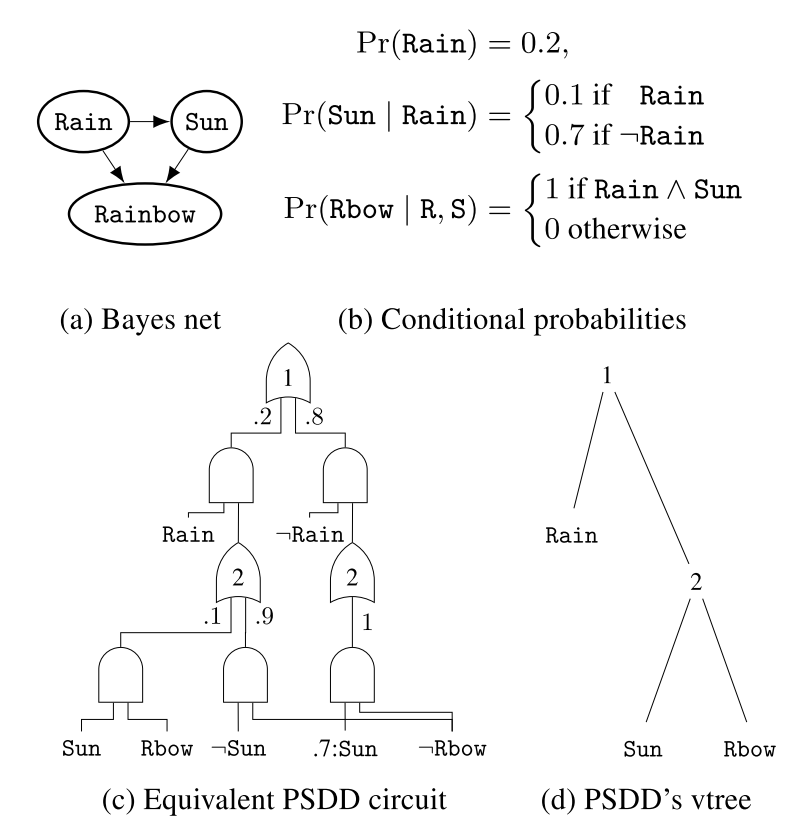}}
\caption{A Bayesian network and its equivalent PSDD ~\cite{liang2017learning}}
  \label{fig:psdd_2}
\end{figure}

Intuitively, PSDDs are 
parametrized directed acyclic graphs (DAGs), as seen in  Figure~\ref{fig:psdd_2}. Here, each terminal node represents a univariate (Bernoulli) distribution over a binary variable (e.g. $B_j$) with a probability $\theta$ represented by the tuple $(\theta_j: B_j)$. Within the tree, each node is either an AND or an OR node. An AND node has two inputs termed prime $p$ for the left one and sub $s$ for the right one. The OR node  can have an arbitrary number of inputs, where each of the $n$ input wires is annotated by a probability $\theta_1,... ,\theta_n$ together making up a normalised distribution over the variables represented by the corresponding \textit{vtree}. Moreover, OR and AND gates always alternate, such that a given OR node can also be represented as a set of AND nodes or decisions: $\{(p_1, s_1, \theta_1),...., (p_n, s_n, \theta_n)\}$.

In order to retain the desirable properties of SDDs for inference (tractability) and canonicity, similar syntactic restrictions hold here as well. Firstly, each of the AND gates has to be \textit{decomposable}, meaning that the \textit{vtree} nodes represented by prime and sub share no variables. In other words, the prime and sub have to represent probability distributions over disjoint sets of variables. Analogously, {\textit{determinism}} demands that for each possible world (or assignment), there can be at most one prime that assigns a non-zero probability to it (the specific world).

In ~\cite{liang2017learning,bekker2015tractable}, a learning regime is proposed  that is capable of learning a PSDD as well as the underlying SDD and \textit{vtree}~\cite{pipatsrisawat2010lower} directly from data. It works by iteratively updating and improving the structure of the PSDD to better fit the data. It does so by applying specified \textit{clone} and \textit{split} operations to a PSDD $r$ at each step. Learning is then carried out until a time limit is reached or a pre-defined \textit{score} converges on the validation data (if present, otherwise training data). This 'score' is based on the log-likelikhood of the model given the data, but takes the size of the tree into account as well.
The log-likelihood of PSDD $r$ given data $\mathcal{D}$ is then a sum of log-likelihood contributions per node:
\begin{equation}
\small
ln\mathcal{L}(r|\mathcal{D}) = lnPr_r(\mathcal{D}) = \sum_{q\in r} \sum_{i \in q} ln \theta_{q,i}\mathcal{D}\#(\gamma_q, [p_{q,i}])
\end{equation}
where $\#(\gamma_q, [p_{q,i}])$ is the number of examples that satisfy the node context of $q$ and the base of $q's$ prime $p_{q,i}$.

Additionally, ~\cite{liang2017learning} also proposed an algorithm for learning ensembles of PSDDs (EM-LearnPSDD) which is built on the learnPSDD algorithm and the soft structural EM algorithm by ~\cite{friedman1998bayesian}. This algorithm consists of two nested learners, where the outer EM is learning the structure and the inner EM is learning the parameters. This is the algorithm we predominantly use in our experiments.

\subsection{Neural Networks \& Autoencoders}
\label{sec:ae}

An AE is a specific instance of an artificial neural network (NN) that is intended to reproduce the given input as an output~\cite{goodfellow2016deep}. It consists of two parts: the encoder \encoder~and decoder \decoder. Internally, it has a hidden layer referred to as the \textit{feature layer} (\fl) such that $\fl= \encoder(x)$ for some input data $x$ and $x_{rec} = \decoder(\fl)$ where $x_{rec}$ is the reconstructed input, $\fl \in \mathbb{R}^{\textit{dim}}$, and $\textit{dim}$ is the dimensionality of the \fl. Restrictions are usually imposed on the structure of the network such as reduced dimentionality in the \fl~or added noise on the input. By reducing the dimension of the \fl~relative to the dimension of the input $x$, the network is intended to learn the
\textit{most valuable} features for reconstructing the original image. The \textit{learning} procedure then constitutes learning the encoder and decoder function simultaneously by minimising the reconstruction loss 
(e.g., mean squared error) penalising $\decoder(\encoder(x))$ for being dissimilar to $x$.

The \textbf{variational autoencoder} (VAE) ~\cite{kingma2013auto,rezende2014stochastic} is a variant of AE that builds on the stochastic generalisation of the classical AE architecture, where instead of a deterministic function, \encoder~and \decoder~are stochastic mappings $p_{\encoder}(\fl | x)$ and $p_{\decoder}(x | \fl)$. Thus, we can also view \encoder~and \decoder~as conditional probability distributions. Utilising this probabilistic interpretation, the VAE framework defines a distribution $q(\fl|x)$ (e.g the Gaussian distribution) such that \fl~samples can be drawn from that distribution. 
Then we can use the Kullback-Leibler divergence ($D_{KL}$) to enforce the encoder network to be as similar as possible to our chosen distribution $q(\fl | x)$ while at the same time maximizing the $\log p(x)$ prior
which is achieved by updating the weights based on the gradient of:
\begin{equation}
\log p(x | \fl) - D_{KL}(q(\fl | x) \| p(\fl))
\end{equation}
The \textit{reparameterization trick} \cite{kingma2013auto} then allows us to enable stochastic gradient descent by reformulating the task using stochastic input layers.

Finally, we leverage the Gumbel-Max trick~\cite{gumbel1954statistical,maddison2014sampling} which yields the 
\textit{Gumbel-Softmax Distribution} that is defined as a continuous distribution over the simplex that can approximate samples from a categorical distribution~\cite{jang2016categorical}.

\section{Methodology \& Evaluation Metrics}
\label{sec:model}
In this section, we propose a novel model for representing 
and learning a symbolic generative model from a neural network that is trained over unstructured data $\mathcal{D}$.
This is possible by means of the intermediate \fl, defined over $n$ discrete variables. The model is to be considered as generative but over a set of \textit{domains}, meaning that we can perform conditional sampling for a given domain with respect to other domains. Domains, written as $D^A,D^B,D^C,\ldots,$ represent disjoint subsets of $\cal D$. For example, suppose we are given an image dataset such as MNIST. Here, the images are denoted by domain $D^A$ (say, $D^A \subset \mathbb{R}^{28\times 28}_{[0,1,...,255]}$) and the corresponding labels by domain $D^Y$ (say, $D^Y \subset \mathbb{N}$)  such that $D^A \sqcup D^Y = \mathcal{D}$. The model is then able to approximate the distributions $p(D^A,D^Y)$ and moreover, can sample from $p(D^A\mid D^Y$) and $p(D^Y\mid D^A)$. 

\subsection{Architecture}

We now discuss the formal and architectural components of our model.

\begin{definition}\textbf{ (Feature layer)}
The \fl~is a finite set of discrete (typically, Boolean) variables; that is, $\fl = \{C_0,C_1,...,C_l\}$ for some $l\geq 0$. Intuitively, \fl~represents an encoded discretized version of the original data $\mathcal{D}$. If the data is split into different domains (disjoint sets, as explained above) $\mathcal{D} = D^A \sqcup \ldots \sqcup D^Y$, then the \fl~can also be split into disjoint domains: $\fl = \fl^A  \sqcup\ldots\sqcup \fl^Y $. The size of \fl~is then the number of variables $|\fl| = l$ and clearly, $|\fl| = |\fl^A|  + \ldots + |\fl^Y|$. Further, for a given domain $\fl^X$, we use $\dom(\fl^X) > 0$ to denote the number of possible values that the discrete variables can take. 
\end{definition}

\subsubsection{Encoders and Decoder Specification}

For a given domain $D^X\in \{D^A,D^B,D^C,..\}$, we have an encoder $\encoder^{X}$ and a decoder $\decoder^{X}$ such that $\fl^{X} = \encoder^{X}(D^X)$ and $d^{X}(\fl^X) = \decoder^X(\encoder^X(D^X)) \subset D^X$. This is like  the usual AE setup, where we map from the input domain (e.g. $D^X$) to an intermediate representation (e.g. $\fl^X$) using the encoder and then map back to the original domain using the decoder. This formulation also works for stochastic encoder/decoder networks with $p_{\encoder}(\fl^X \mid D^X)$ and $p_{\decoder}(D^X \mid \fl^X)$ utilising the Gumbel-Softmax distribution.

Essentially, encoders and decoders are functions of varying complexity depending on the domain. While some encoder-decoder pairs may be learned from data, others can be defined deterministically. Revisiting the MNIST dataset, for example, here we may define domain $D^A$ to be the images (e.g. $a \in D^A \subset \mathbb{R}_{[0,1,..,255]}^{28 \text{ x } 28}$) and domain $D^Y$ to represent the corresponding labels (e.g. $y \in D^Y = \{0,1,..,9\}$). In particular, the encoder and decoder for domain $D^A$ ($\encoder^A, \decoder^A$) are deep convolutional neural networks. As for domain $D^Y$, the encoder/decoder can be defined as the \textit{one-hot} encoding and the $\argmax$ function repectively ($\encoder^Y(y) \mapsto one\_hot(y)$, $\decoder^Y({fl}^Y_i) \mapsto \argmax_j(fl^Y_{ij})$).

The learning of these functions (if applicable) is done in an unsupervised manner using VAEs in our setup and is referred to as $\textit{learning phase I}$ throughout this article. A visualization of the pipeline can be seen in Figure~\ref{fig:phase_2}.

\subsubsection{The Logical Interpretation}

The (logical) generative model represents the dependencies between the individual variables of the \fl; in other words, it represents the joint probability distribution over the variables of the \fl, i.e., $Pr(\fl)$. In this paper, we chose to use PSDDs~\cite{kisa2014probabilistic} though other models such as sum product networks (SPNs) ~\cite{poon2011sum} could have been used as well. 
This choice was based on the reported ability of PSDDs to handle constraints in the learning regime. These could be one-of label constraints or any other kind of Boolean function over the inputs. 


\subsubsection{Learning}

\begin{figure}
    \centering 
    \includegraphics[width=.9\columnwidth]{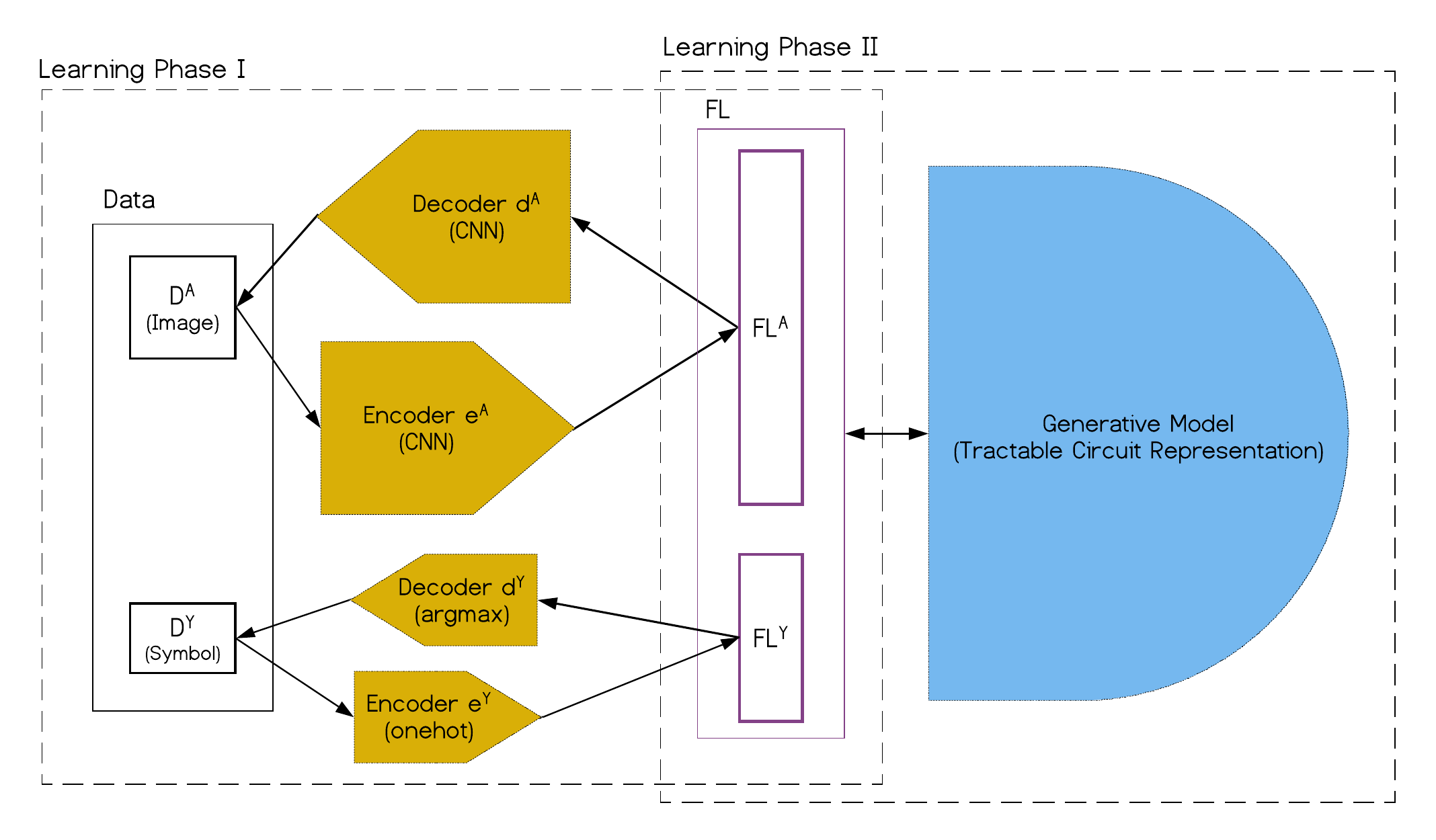}
    \caption{The two phases of the proposed learning system w.r.t. the MNIST example}
    \label{fig:phase_2}
\end{figure}

The learning of the system is done in two phases (see Figure~\ref{fig:phase_2}). Learning phase I denotes the learning of the encoder-decoder function pairs for each domain (independently and unsupervised). Once learning phase I is completed we can use the encoders to map the data to the \fl~representation and learn the logical generative model (learning phase II), that is, the PSDD. Essentially, the variables of the \fl~are the propositions of the PSDD.

\subsection{Querying}

Due to the generative property of PSDDs, we are able to perform any query of the form: $Pr(q|v) = \frac{Pr(q \land v)}{Pr(v)}$ where $q$ is the query and $v$ is the evidence, both Boolean functions over the variables of the \fl. Furthermore, by Theorem 7 of ~\cite{kisa2014probabilistic}  such probabilities can be computed in one pass through the tree and thus in polynomial time w.r.t. the size of the graph. (Thus, these are referred to as tractable models in the literature.) Specifically, in the MNIST example, such queries would take the form: 
$Pr(e^X(q)\mid e^Y(v))$ where $X$ and $Y$ correspond to two domains of the data ($D^X, D^Y \subset \mathcal{D}$ and $q \in D^X, v \in D^Y$).

\subsubsection{Generative Query}
\label{sec:genquery}

Given evidence $v$, we define the task of a \textit{generative query} as one that samples values for all variables in the \fl~which are not assigned in the evidence. That is, $fl = generativeQuery(\Delta, v)$, which is discussed in Algorithm~\ref{alg:gernativequery} and is equivalent to $fl \sim Pr(\fl \mid v)$.

As mentioned before the resulting assignments of variables returned from the algorithm can then be decoded using the decoder \decoder.

\begin{algorithm}
\caption{generativeQuery(PSDD $\Delta$, evidence $v$, categorical dimension $k$)}
\label{alg:gernativequery}
\begin{algorithmic}[1]
    \STATE $\textit{assigned} \gets \text{variables\_appearing\_in}(\textit{v})$
    \STATE $\textit{not\_assigned} \gets \text{variables\_appearing\_in}(\Delta) - assigned$
    \STATE $generated \gets dict()$
    \WHILE{not empty(not\_assigned)}
        \STATE $var \gets pop\_random(not\_assigned)$
        \STATE $dist \gets zeros(dim = k)$
        \FOR{$j \in range(k)$}
            \STATE $dist_j \gets \frac{Pr_{\Delta}(var = onehot(j)|v)}{Pr_{\Delta}(v)}$
        \ENDFOR
        
        \COMMENT{Sample from the categorical distribution}
        \STATE $inst \gets sample(dist)$ 
        
        \COMMENT{Add sampled assignment to $v$}
        \STATE $v \gets (v \land (var = inst))$
        
        \COMMENT{Add sampled assignment to output dictionary}
        \STATE $generated[var] \gets inst$
    \ENDWHILE
\RETURN $generated$    
\end{algorithmic}
\end{algorithm}

\subsection{Evaluation}
\label{sec:evaluation}

In order to evaluate our model on image datasets, we focus on two main aspects. 
First, classification accuracy and secondly, the \textit{recoverability} of the trained model in terms of how it \textit{interprets} a given domain, explained below.  

Classification accuracy is used as a quantifiable score that is easily comparable to other learning systems. We train the model on multiple image datasets before asking the model to classify unseen images ($a \in D^A$) into one of the possible categories ($y \in D^Y$) using the following formulation (here $'$ denotes a sample drawn from a distribution):
\begin{align}
        y' &= \decoder^Y(gernerativeQuery(\Delta, \encoder^A(a)))\\
        \encoder^Y(y') &\sim Pr(\fl^Y \mid \encoder^A(a))\\
        y' &= \decoder^Y(fl'^{Y} \sim Pr(\fl^Y \mid [fl'^{A} \sim p_{\encoder}(\fl^A \mid a) ]))
\end{align}

We investigate the interpretability of the model by manually inspecting samples drawn from the distribution for some evidence. For the MNIST dataset, for example, we sample images for each category or class and check if the images correspond to the class. Such samples will be computed as follows (where $a\in D^A$ denotes the images as before, and $y \in D^Y$ are the class labels):
    \begin{align}
        a' &= \decoder^A(generativeQuery(\Delta, \encoder^Y(y))\\
        \encoder^A(a') &\sim Pr(\fl^A \mid \encoder^Y(y))\\
        a' &\sim p_{\decoder}(D^A \mid [fl'^a \sim Pr(\fl^A \mid \encoder^Y(y))])
    \end{align}
    
Finally, we analyse the  variables of the  \fl. This sheds some light on the inner workings of the model, and gives us an insight into what the individual variables capture. Basically, we approximate the expectation of decoded \fl~samples where, in a binary setting, samples would be drawn conditional on a specific variable being true or false:
\begin{equation}
\begin{split}
    \textit{diff}_{\fl_i} = \mathbb{E}_{fl' \sim p(\fl \mid fl_i)} \decoder^A(fl')\\ - \mathbb{E}_{fl' \sim p(\fl \mid \neg fl_i)} \decoder^A(fl')
\end{split}
\end{equation}
This is approximated by: 
\begin{equation}
\begin{split}
\approx \frac{1}{N}*\sum_{fl' \sim p(\fl \mid fl_i)}^N \decoder^A(fl') * p(fl')\\ -\frac{1}{N}*\sum_{fl' \sim p(\fl \mid \neg fl_i)} \decoder^A(fl') * p(fl')
\end{split}
\end{equation}
Here we define the decoded image to be $w$ pixels in width and $h$ pixels in height. Then each greyscale image $a$ is normalized to be an element of $a \in [0,1]^{w * h}$. What follows is that $\textit{diff}_{\fl_i} \in [-1,1]^{w * h}$ and as such it has to be normalized accordingly in order to produce an image.
\begin{align}
    \label{eq:visual_fl}
    \textit{visual}_{\fl_i} &= \Bigg[\frac{\textit{diff}_{\fl_i} + 1}{2},\frac{-\textit{diff}_{\fl_i} + 1}{2}\Bigg]
\end{align}

Here, $\textit{visual}_{\fl_i}$ is depicted as a tuple of images corresponding to the variable $i$ being true vs. false and vice versa.

\section{Experiments}

In this section, we investigate the predictive accuracy on unseen data (via  a held out test set) as well as the generative power of the model. 
Firstly, we consider the standard classification task. Secondly, we run experiments where noise is added to the label of each entry; that is, each training entry has $k$ additional random labels specified (in addition to the correct one). Thirdly, we explore tasks which consist of at least two images and possibly a symbolic value. In one such experiment for example, a data point is defined over two images, representing successive integers and we are then interested in generating one image given the other (e.g., generate \textit{7} if the first image is \textit{6}.). These are referred to as \textit{functional tasks}. We conclude with an analysis of the \fl. A comprehensive listing of the experimental setups and experiments run are given in the supplement Appendix.

\subsubsection{Data} In order to get a comprehensive understanding of the capabilities of the proposed model, we used three different datasets. First, the MNIST dataset~\cite{lecun1998gradient} containing $10^5$ 28x28 (grayscale) images that represent handwritten digits, along with the corresponding class label. After the first set of experiments on the MNIST dataset, we used the hyper-parameters for the best performing models and re-run the experiments on the FASHION dataset~\cite{xiao2017fashion}, which contains $10^4$ 28x28 (grayscale) images of fashion items and the corresponding labels belong to one of the 10 categories. Finally, to investigate the scalability of the model, we used the EMNIST (extended-MNIST)~\cite{cohen2017emnist} dataset, where the images are handwritten numbers and letters of the English alphabet with the corresponding labels (47 classes in total).

\subsubsection{Hardware} Since most experiments involved two training phases using very different optimization methods, we made use of two different cluster architectures in order to improve performance. Learning phase I is concerned with learning the parameters of a deep NN using mini-batch gradient descent, and back propagation were run on GPU clusters. The cluster nodes used here are a combination of Dell PowerEdge R730 and Dell PowerEdge T630. Each has two 16 core Xeon CPUs, where the GPUs use NVIDIA cards Tesla K40m, GeForce GTX Titan X and GeForce Titan X. 
Learning phase II, on the other hand, uses the learnPSDD structure learning algorithm, and this was run on CPU clusters, where each of the 21 nodes is a Dell PowerEdge R815 with four 16 core Opteron CPUs and 256GB of memory.

\subsection{Classification Task}
Given training examples consisting of images and their labels, we trained the encoder-decoder pair unsupervised in the first instance, and the PSDD on the whole \fl~($\fl^A$ + $\fl^Y$ representing the image and label respectively) in the second instance. The hyper-parameters explored here include the \textit{vtree}-search algorithm used, as well as the option of compressing the label, a one-hot encoding to a binary representation (e.g. $[0,1,0,0] \mapsto [1,0]$). Furthermore, we varied the number of variables in $\fl^A$ and the categorical dimension of such variables (denoted by $|\fl^A|$ and $\dom( \fl^A)$ respectively). Note that the categorical dimension essentially corresponds to whether we interpret the features in a Boolean space vs finite multi-valued space.

\subsubsection{MNIST}
\begin{figure}
    \centering 
    \includegraphics[width=1\columnwidth]{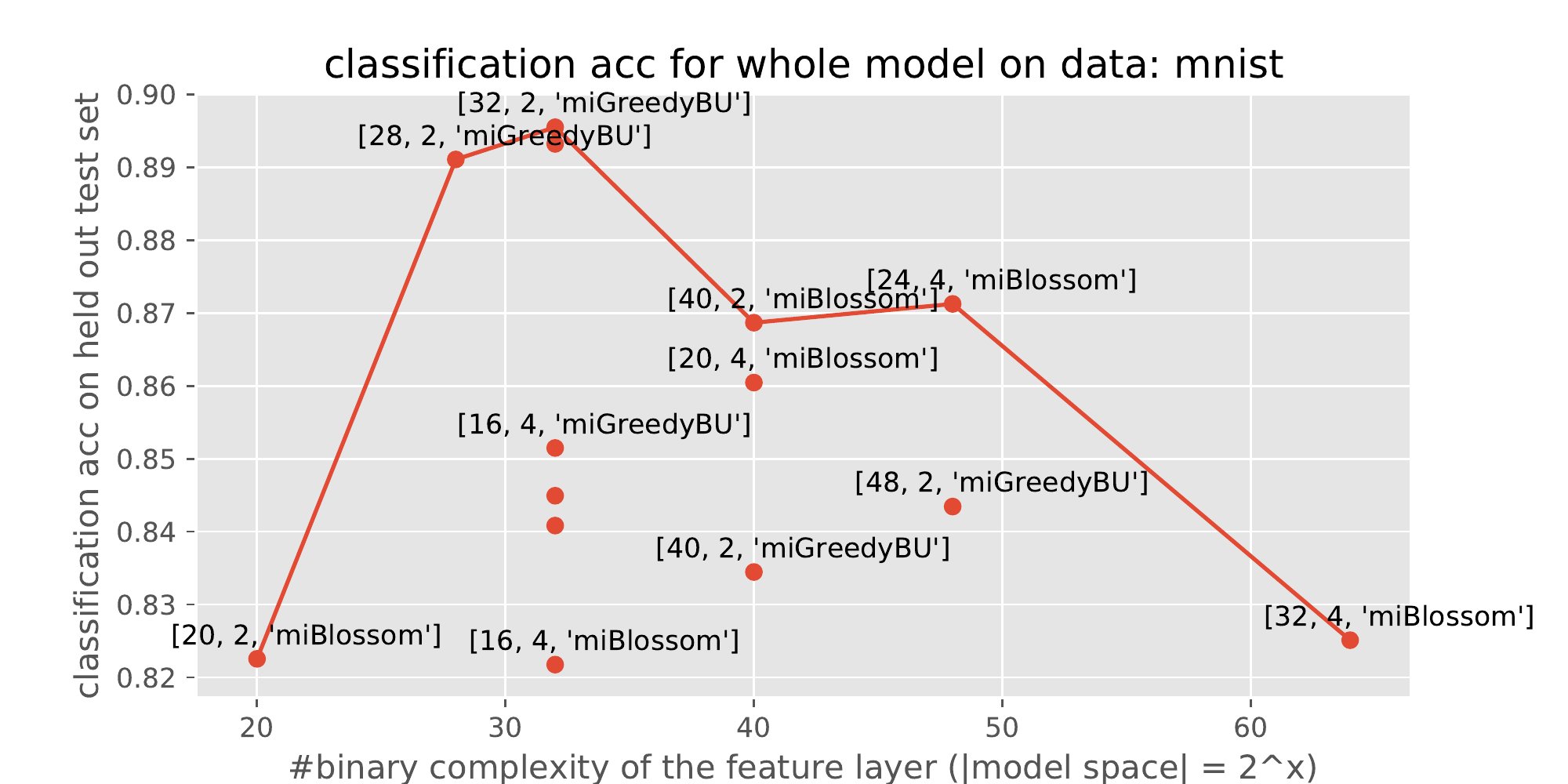}
    \caption{Experiment classification results MNIST (label format: $[| \fl^A |,\dom( \fl^A)$, vtree seach algorithm)}
    \label{fig:var_mnsit_losses_graph}
\end{figure}

The best classification accuracy on the MNIST dataset was measured at: $89.55\%$ using 32 binary variables  (and a categorical dimension of 2) and a one-hot encoded $\fl^Y$. In comparison, we note that discriminative models such as convolutional NN achieve 99.3\%~\cite{lecun1998gradient} on the same task. A more comprehensive overview is given in Figure~\ref{fig:var_mnsit_losses_graph}. Here we can  see that there is a clear trade off between expressiveness of the \fl~and the ability for the PSDD learning to interpret this \fl. In other words, if the \fl~is too small, then the neural model will not be able to learn a meaningful mapping, retaining valuable information in the encoding, and if the \fl~is too large, the PSDD learner struggles to find correlations between the variables. 

\begin{figure}
\centering
    \begin{subfigure}[t]{.49\columnwidth}
            \centering 
            \includegraphics[width=.9\columnwidth]{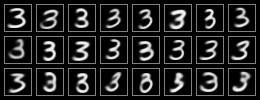}
            \caption{class:3 (MNIST)}
            \label{fig:class_mnist_3}
    \end{subfigure}
    \begin{subfigure}[t]{.49\columnwidth}
        \centering 
        \includegraphics[width=.9\columnwidth]{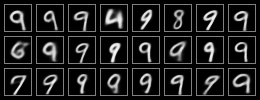}
        \caption{class:9 (MNIST)}
        \label{fig:class_mnist_9}
    \end{subfigure}
    \begin{subfigure}[t]{.49\columnwidth}
        \centering 
        \includegraphics[width=.9\columnwidth]{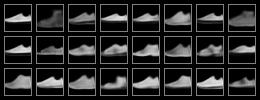}
        \caption{class:Sneaker (FASHION)}
        \label{fig:class_fashion_3}
    \end{subfigure}
    \begin{subfigure}[t]{.49\columnwidth}
        \centering 
        \includegraphics[width=.9\columnwidth]{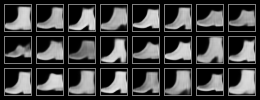}
        \caption{class:Ankle boot (FASHION)}
        \label{fig:class_fashion_9}
    \end{subfigure}
\caption{Sampled images for specified classes of models trained on classification task (MNIST and FASHION) with $|\fl^A | =32$, $ \dom(\fl^A) =2$}
\label{fig:classification_sampels}
\end{figure}

From the best setup above, to test the generative abilities, we sampled images for each of the 10 categories using the proposed conditional-sampling algorithm. These samples are depicted in Figures~\ref{fig:class_mnist_3},~\ref{fig:class_mnist_9} for 2 of the 10 classes. Since these are samples, we should expect to see some variation, corresponding to the Figures. In a sense, the system demonstrates a prototypical understanding of what the labels  represent. It is interesting to relate this insight to approaches such as~\cite{lake2015human} that involve an explicit token construction framework for generating images. We imagine that it might be possible to use the variables induced in our frameworks as a token generator, which we leave for the future.

\subsubsection{FASHION}
For the FASHION dataset, we achieve a classification accuracy of $75\%$ using the hyperparameters of the best performing MNIST model. Interestingly enough, the reconstruction loss (binary-cross-entropy) of the neural model is smaller (thus, better) in this scenario than in the MNSIT case, and the PSDD score is larger (thus, better) as well. The predictive accuracy on the held out test set is still considerably lower. This can be due to many reasons, most notably perhaps due to the additional complexity of the images. 
The additional variability of FASHION influences the computed reconstruction loss of the VAE, as it computes an average over pixel difference between the original image and the reconstructed one. For testing the generative abilities, once again, we sampled images for 2 of the 10 classes, as shown in Figures~\ref{fig:class_fashion_3} and \ref{fig:class_fashion_9}.

\subsubsection{EMNIST}
Finally, when running experiments on the EMNIST dataset, we found that the system is not quite capable of scaling to such a large number of image classes, which we suspect seriously affects the performance of the PSDD learner. Additionally, the VAE is confronted with a much more complex task in differentiating symbols (e.g., ``1'' and ``l''). Here we only recorded an overall best accuracy of $29\%$, where $1/47 = 0.021$ would be the expected random accuracy. It is an interesting question for the future to consider how to handle so many image classes with a Boolean learner.

\subsection{Noisy Label Task}
As mentioned earlier, we are interested in challenging the system by providing  $k$ randomly generated additional labels to the correct one during training.  To evaluate the experiment, we computed the accuracy on a held out set (for MNIST) only containing the right label (no noise). 
Generally speaking, as expected, we observed a decreasing accuracy with increasing noise: for example,  adding one additional label (noisy-1) decreases accuracy by $.05$ to $85.0\%$. Even if 2 and 3 noise labels are added, the accuracy only decreases to $82.1\%$ and $71.8\%$ respectively. Intuitively, the experiment requires the PSDD to reason about the possible labels for a given image and thus, we show that the logical model performs this reasoning in a satisfactory manner. 

\subsection{Functional Tasks}

The idea here is to have training examples consisting of at least two images and maybe a symbolic value that denotes the relationship between these images. 
However, no semantic characterisation is provided for this symbolic value in our setup, so the system tries to map the image pairs to the value purely from visual features. (Thus, the machinery of logically defining such functions, as seen in, e.g., DeepProbLog~\cite{manhaeve2018deepproblog}, could be used to extend our framework further.)

\begin{figure}
\centering
    \begin{subfigure}{.23\textwidth}
        \centering 
        \includegraphics[width=1\columnwidth]{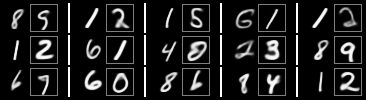}
        \caption{Sampled images for $\fl^A$}
        \label{fig:exp1_con_succ_a}
    \end{subfigure}
    \begin{subfigure}{.23\textwidth}
        \centering 
        \includegraphics[width=1\columnwidth]{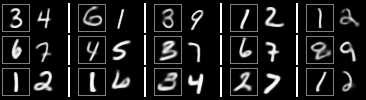}
        \caption{Sampled images for $\fl^B$}
        \label{fig:exp1_con_succ_b}
    \end{subfigure}
\caption{Image generation for successor task, one datapoint is a tuple, where the image with border was sampled for image with no border}
\label{fig:exp1_conditional_succ}
\end{figure}

The simplest one is where we provide an image, and expect the system to generate a second image such that the integers present in the images are successors. This demonstration is depicted  in Figure~\ref{fig:exp1_conditional_succ}, where we observe that the predecessor/successor integer's image was generated successfully. 

\begin{figure}
\centering
    \begin{subfigure}[t]{.23\textwidth}
            \centering 
            \includegraphics[width=1\columnwidth]{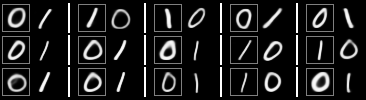}
            \caption{(MNIST) Sampled images (with border) for $\flinst^Y = 1$}
    \end{subfigure}
    \begin{subfigure}[t]{.23\textwidth}
        \centering 
        \includegraphics[width=1\columnwidth]{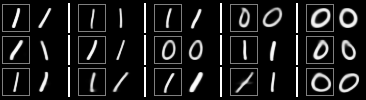}
        \caption{(MNIST) Sampled images (with border) for $\flinst^Y=0$}
    \end{subfigure}
    \begin{subfigure}[t]{.23\textwidth}
            \centering 
            \includegraphics[width=1\columnwidth]{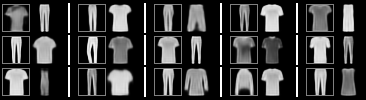}
            \caption{(FASHION) Sampled images (with border) for $\flinst^Y=1$}
    \end{subfigure}
    \begin{subfigure}[t]{.23\textwidth}
        \centering 
        \includegraphics[width=1\columnwidth]{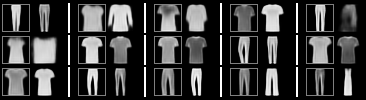}
        \caption{(FASHION) Sampled images (with border) for $\flinst^Y=0$}
    \end{subfigure}
\caption{Image generation for binary-logic-\textit{XOR} task, one datapoint is a tuple, where the image with border was sampled for image with no border and $\flinst^Y \in \{0,1\}$}
\label{fig:ex_rel_blxor}
\end{figure}

In an additional set of experiments, we also provide a symbolic variable ($\fl^Y$) that is the evaluation of a mathematical function over the two images.
One example of such a function is the Boolean logic \textit{XOR}. Here, we first train the unsupervised VAE on the whole (e.g., MNIST) data, and then create a custom training dataset where each entry contains two images, either ``0" or ``1", and the result of applying the logical \textit{XOR} operation on the label of these two images ($\fl^Y$). Thus, the \fl~in this task is made of three individual parts: $\fl^A = \encoder(\textit{imgA}), \fl^B = \encoder(\textit{imgB})$ representing two images and $\fl^Y = \textit{bool}(\textit{label}(\textit{imgA}))\textit{ XOR } \textit{bool}(\textit{label}(\textit{imgB}))$, representing the evaluation of the \textit{XOR} function on the original labels of the two images. We can then evaluate the accuracy on the correctness of the predicted symbolic value on a held out test set. Conversely, we can sample for one of the images given the other image and a specified $\fl^Y$ value. To reiterate, this is purely visual reasoning, so to clarify that, we can also repeat the experiment with the FASHION dataset, treating T-shirts and Trousers to correspond to \textit{true} and \textit{false} respectively~(e.g., $\textit{bool}(\text{Trouser})=1$). (All other digits in MNIST and all other image classes in FASHION are discarded.) The classification accuracy that we measured on a held out test set were $99.4\%$ and $88.2\%$ for MNIST and FASHION respectively. Generated samples are shown in Figure~\ref{fig:ex_rel_blxor}. As an example of a more complex function, we also conducted experiments on MNIST, where $\fl^A$ and $\fl^B$ range over all images present in the dataset 
but $\fl^Y$ constitutes the result of the arithmetic \textit{plus} operation on the original labels of the two images, such that $\fl^Y = \textit{label}(\textit{imgA}) + \textit{label}(\textit{imgB})$. Here, we have many more possible $\fl^Y$ values and multiple combinations of images that correspond to the same $\fl^Y$ value. However, the recorded classification accuracy on the held out test set is only $9.92\%$. Thus, the conclusion to be drawn here is although the logical generative model does allow us to formulate challenging tasks over mathematical and logical functions, it currently only resolves this in terms of the visual features. So a second interesting direction for the future is to understand how to go beyond this and find a way to incorporate (or learn) the semantic meaning of the mathematical function. PSDDs~\cite{liang2017learning}, for example, can be trained with constraints which might offer a possible way to make progress in this direction. 

\subsection{\fl~Analysis}

\begin{figure}
\centering
    \begin{subfigure}[t]{.23\textwidth}
            \centering 
            \includegraphics[width=.9\columnwidth]{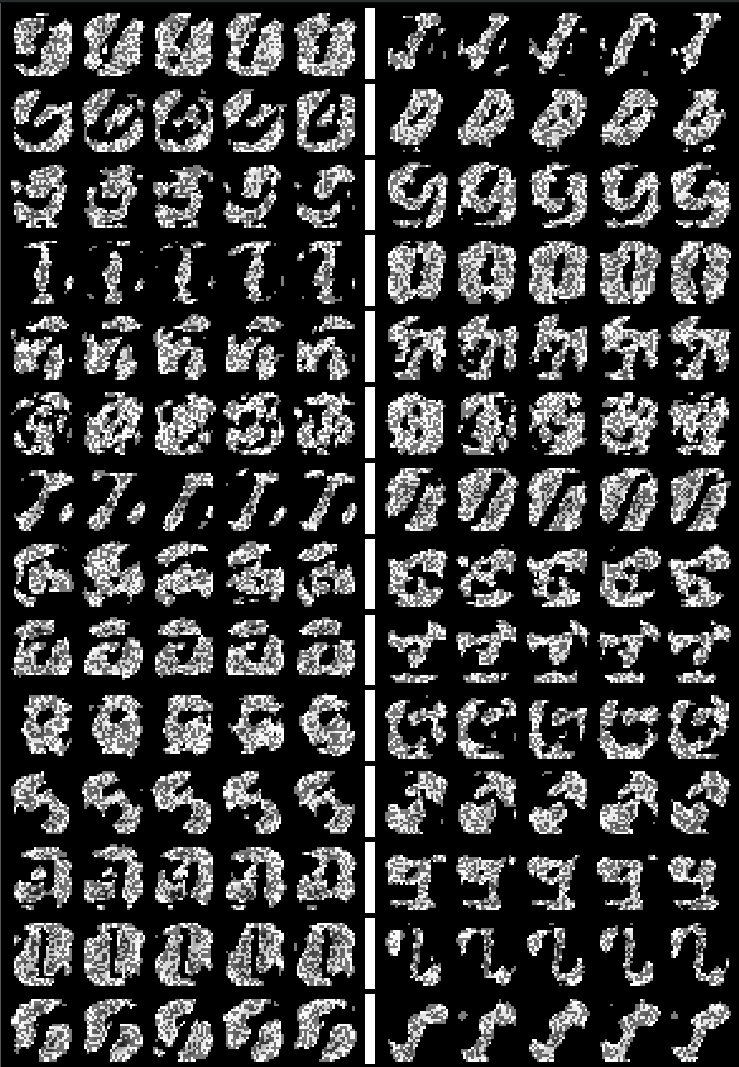}
            \caption{$visual_{fl_i}$ on MNIST}
            \label{fig:fl_analysis_mnist}
    \end{subfigure}
    \begin{subfigure}[t]{.23\textwidth}
        \centering 
        \includegraphics[width=.9\columnwidth]{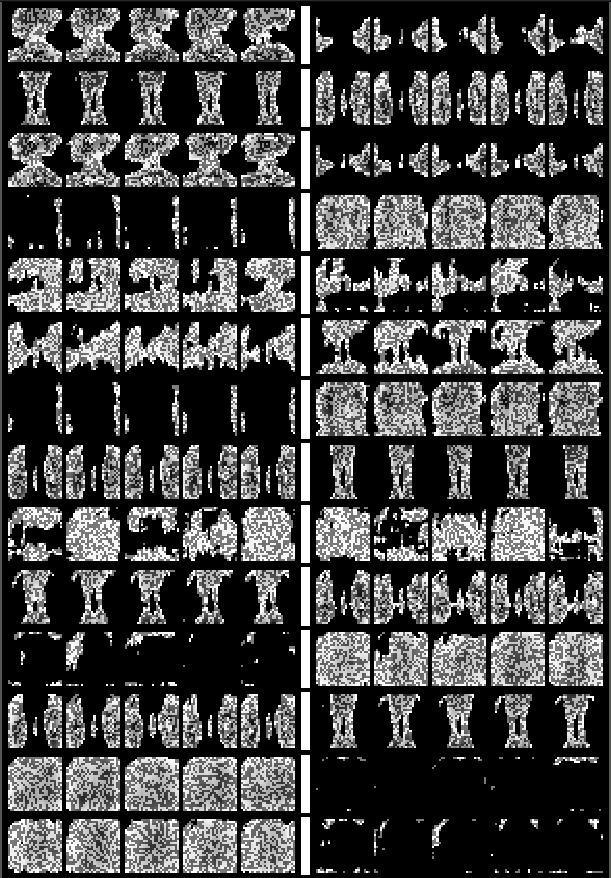}
        \caption{$visual_{fl_i}$ on FASHION}
    \end{subfigure}
\caption{Visualisation of the first 14 binary variables in the classification models for two datasets (variables are ordered)}
\label{fig:fl_analysis}
\end{figure}

To understand what a given variable in the \fl~actually represents, we use the generative query algorithm and Equation~\ref{eq:visual_fl}. To evaluate this more concretely, we sample images for our best performing model on MNIST and FASHION. In Figure~\ref{fig:fl_analysis}, we computed $visual_{\fl_i}$ five times with $N = 200$ for each variable of $\fl^A$. In Figure~\ref{fig:fl_analysis}, each row corresponds to one of the first 14 variables of the \fl~(in order from 1 to 14, top to bottom). What this demonstrates is that the model accords meaningful elements of the images to each variable of the \fl. Indeed, we see that individual variables correspond to different shapes such as slightly bent lines  in row/variable 1 of Figure~\ref{fig:fl_analysis_mnist} or circular objects  in row/variable 4. In a sense, the \fl\ is able to identify discrete visual components for the images, which hints at a compact and compositional understanding of the domain in question.

\section{Conclusion \& Discussion}

In this work, we were interested in understanding what precisely the latent space of AEs capture, and whether that space could be inspected from a logical viewpoint. In that regard, we motivated the learning of a symbolic generative model on the \fl, which allows us to inspect the hidden layers and perform logical reasoning over the variables of these layers. For example, by means of a conditional sampling algorithm, we were able to generate prototypical images for a label, and moreover, generate labels for images.

As mentioned previously, with regards to the standard classification task we see that our model can not compete with other state-of-the-art systems such as deep convolutional neural networks (CNNs). However, when making such comparisons, one should consider that discriminate models such as CNNs are not generative, whereas 
generative models (e.g. VAEs) are not appropriate for  classifying 
images as they may not discriminate between individual images. 

Although one might, in general, consider that a lower  performance is a reasonable trade-off in exchange for increased functionality and interpretability, we do not think this is ``fundamental" tradeoff:  our observation has been that the PSDD software seems more capable of handling intricate Boolean reasoning in comparison to SPN's, but both struggle somewhat when considering multi-valued discrete variables. Note that by increasing the ``encoding” space, we are allowing for more granular reconstructions of the latent space, and so we should expect to reach the performance of state-of-the-art models. However, unfortunately, the PSDD software struggles when considering many   multi-valued discrete variables, so there is an engineering effort required. In contrast, many conventional deep learning software packages have benefited from considerable optimizations. 

In addition to classifying images, the model was put to test in challenging tasks capturing structural, logical or mathematical relationships between pairs of images, as well as the handling of noisy labels. While we did observe scalability issues when considering a very large set of classes, the underlying framework still offers an insightful logical view of the hidden layers. This provides the space for interesting avenues for the future, such as integrating our framework with existing neuro-symbolic frameworks. In particular, can one of these frameworks provide a way to reason about mathematical functions in a semantic manner (perhaps also learn them),  rather than the purely visual quality exploited in the current setup? Can proposals from statistical relational learning~\cite{getoor2007statistical} help us capture and reason about intricate logical relationships between variables? The overall goal, then, is to get a better grasp of how abstract concepts and high-level reasoning might emerge from low-level sensory data. We hope that this work, which attempts to re-purpose a deep learning framework in logical space, provides some of the insights on how that is possible, and at the same time, shows the benefits of using a symbolic generative model in a differential latent space. 

\section*{Acknowledgements}
We would like to thank John Quinn for his valuable input on this work and many fruitful conversations.
Anton Fuxjaeger was supported by the Engineering and Physical Sciences Research Council (EPSRC) 
Centre for Doctoral Training in Pervasive Parallelism (grant EP/L01503X/1) at the University of Edinburgh, School of Informatics. Vaishak Belle was supported by the Royal Society University Research Fellowship. We would also like to thank our reviewers for their helpful suggestions.

\bibliographystyle{apalike}
\bibliography{logical_interpretations_of_autoencoders.bib}


\end{document}